# Language Model-Driven Unsupervised Neural Machine Translation

Wei Zhang[1*], Youyuan Lin[1*], Ruoran Ren[1], Xiaodong Wang[2], Zhenshuang Liang[2], Zhen Huang[2]

1. Ocean University of China，
2. Global Tone Communication Technology Co., Ltd.(Qingdao)

**Abstract**

Unsupervised neural machine translation(NMT) is as- sociated with noise and errors in synthetic data when executing vanilla back-translations. Here, we explic- itly exploits language model(LM) to drive construction of an unsupervised NMT system. This features two steps. First, we initialize NMT models using synthetic data generated via temporary statistical machine trans- lation(SMT). Second, unlike vanilla back-translation, we formulate a weight function, that scores synthetic data at each step of subsequent iterative training; this allows unsupervised training to an improved outcome. We present the detailed mathematical construction of our method. Experimental WMT2014 English-French, and WMT2016 English-German and English-Russian translation tasks revealed that our method outperforms the best prior systems by more than 3 BLEU points.

# Introduction

Neural machine translation (NMT) has made remark- able progress in recent years(Sutskever, Vinyals, and Le, 2014; Cho et al., 2014; Bahdanau, Cho, and Bengio, 2014; Vaswani et al., 2017). However, NMT systems exploit many parallel data, and perform less well than statistical machine translation(SMT) systems under resource-poor conditions(Koehn and Knowles, 2017). Thus, NMT op- timization for resource-poor environments has attracted a great deal of interest. Parallel corpora are costly, and may be resource-poor in terms of language pairs. Ef- forts are underway to use the more readily available monolingual corpora to improve NMT systems.

One of the most effective methods is back-translation (Sennrich, Haddow, and Birch, 2015); a source-to-target translation system is trained using synthetic corpora gen- erated by a backward model. Iterative back-translation is also promising (Zhang et al., 2018; Hoang et al., 2018). Language models (LMs) may be of assistance. In the context of unsupervised NMT, some authors (Artetxe et al., 2017; Lample et al., 2017, 2018) have leveraged LMs by training a seq2seq system (Sutskever, Vinyals, and Le, 2014) to serve as a denoising auto-encoder (DAE) (Vincent et al., 2008). Finally, initialization is also of con- cern in the context of resource-poor NMT. Cross-lingual lexica derived from monolingual corpora are widely used to initialize unsupervised NMT systems (Artetxe et al., 2017; Lample et al., 2017). In summary, as Lample et al. (2018) have noted, research on resource-poor NMT focuses principally on: 1) back-translation; 2) use of an LM; and, 3) initialization.

Here, we engage in LM-driven unsupervised construc- tion of an NMT system. Given source sentences, we aimed to estimate accurately the posterior distributions of target sentences. If resources are poor, we compro- mise; we train the NMT system (in an unsupervised manner) to estimate the marginal distributions of target sentences. We derive a weight function for synthetic data based on well-trained LMs and a translation model. However, given the lack of correction during training, convergence of an unsupervised NMT system depends heavily on the initial parameters. Therefore, we use data generated by an unsupervised SMT constructed with the aid of an LM, and cross-lingual embedding, to jump- start training without modifying the NMT architecture. Figure 1 shows the training process.

Experiments using the WMT2014 and WMT2016 datasets showed that our unsupervised NMT system was comparable to that with the optimal baseline (Lam- ple et al., 2018) in terms of English-French tasks, and about 3 BLEU better on English-German and English- Russian tasks; we have raised the bar of state-of-the-art performance.

[*] Equal contribution, Wei Zhang: weizhang@ouc.edu.cn, Youyuan Lin: linyouyuan@stu.ouc.edu.cn

Our contributions are:

1. We show how an LM can drive construction of an unsupervised NMT system. Then, we use a weight function to correct training without changing the NMT architecture; this is simple but effective.
2. We explore how the initial synthetic data influence convergence during training, and we then use an SMT method to boost the quality of initial synthetic data. This is simple, rapid, and requires only off-the-shelf software.
3. We test the system using English-German, English- French, and English-Russian language pairs; our method is the best currently available.

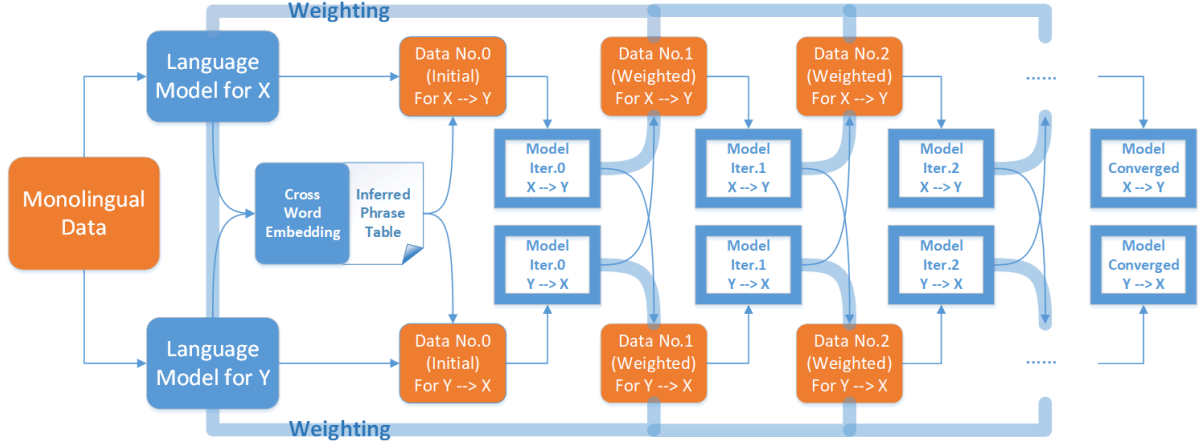

Figure 1: Framework of Language Model Driven Unsupervised Neural Machine Translation. Monolingual corpus of language *X* and *Y* are given. As prepared, we train two LMs, include their cross-lingual embeddings, with which we infer a phrase table. Based on these materials, data for initialization will be generated. Training process will jump-start whereby these initial synthetic data. After initialization iterative back-translation starts, we still use a backward model to sample candidate synthetic data. Different from vanilla back-translation, we weight synthetic data by translation model and both two LMs(indicated by thick blue lines).

# Background

## Iterative Back-translation for NMT

NMT is currently favored. NMT features an attention- base encoder-decoder structure within a recurrent neural network (Bahdanau, Cho, and Bengio, 2014) or a transformer (Vaswani et al., 2017). Given a parallel corpus $\{(x^{(n)}, y^{(n)})\}_{n=1}^{N}$, where N,$(y^{(n)}, x^{(n)})$ denote the the size of the corpus and each pair of parallel sentences (respectively), an x → y NMT model directly maximize the conditional log-probability associated with the parameter θ, as follows:

$$\mathcal{L}_{para}(\theta) = \frac{1}{N}\sum_{n=1}^{N} \log P(Y^{(n)}|X^{(n)};\theta) \qquad (1)$$

Training objective 1 cannot be achieved when only monolingual corpora $\{y^{(n)}\}_{n=1}^{N}$ are available. Instead, iterative back-translation is used to sample x and then maximize the likelihood of the synthetic data (Zhang et al., 2018):

$$\mathcal{L}_{mono}(\theta) = \frac{1}{N}\sum_{n=1}^{N}\sum_{X} P(X|Y^{(n)})\log P(Y^{(n)}|X;\theta) \qquad (2)$$

Such iterative back-translation uses a backward model P(x|y; θback) to estimate the real posterior distribution P(x|y (n) ); however, noise is introduced.

## Cross-lingual Word Embedding

A word embedding is a continuous representation of words. Cross-lingual word embeddings share vector spaces across multiple languages, and are usually trained by deriving a rotation matrix M that maps source embedding onto target embedding (Conneau et al., 2017). Thus, the distances between cross-language embeddings can be calculated; these reveal candidate word-level translations. The translation probability from word xi to yj is:

$$P(y_i|x_i) = \frac{\exp \lambda \cos < e_{x_i}, e_{y_j} >}{\sum_y \exp \lambda \cos < e_{x_i}, e_y >} \quad (3)$$

where ew is the cross-lingual embedding of word w, and λ is a hyper-parameter controlling the peakiness of the distribution. We use the training/inferential methods of Artetxe, Labaka, and Agirre (2018) .

# Framework

## Overview

Figure 1 illustrates the training flow, corresponding algorithm 1. For Language X and Y , we first train two LMs using large amounts of monolingual data. We then train the cross-lingual word embeddings and develop a phrase table. Next, we generate initial synthetic data using Eqs.17 and 18 and use the data to initialize the NMT system. We then commence iterative joint training; at each step, we weight the synthetic data for both LMs as indicated by Eq. 14.

---

**Algorithm 1:** LM Driven US-NMT

**Input:** Monolingual Corpus $M_x$ for language X; Monolingual Corpus $M_y$ for language Y.

**Output:** $\overrightarrow{\theta}$ , parameters of **x** → **y** NMT system; $\overleftarrow{\theta}$ , parameters of **y** → **x** NMT system.

Train $LM_x$, $LM_y$, Word Embedding $E_x$, $E_y$ on $M_x$, $M_y$; Train Cross-lingual Embeddings $E_{cross}$ on $E_x$, $E_y$;

Infer Phrase Table T on $E_{cross}$;

Epoch:=0;

Randomly initialize $\overrightarrow{\theta}$ , $\overleftarrow{\theta}$ ;

**while** *Not converge* **do**

    Randomly select sub dataset $D_y$,$D_x$ from $M_y$, $M_x$;

    **if** *Epoch = 0* **then**

        Generate pseudo sentences $F_x$, $F_y$ by $LM_x$, $LM_y$, T on $D_y$, $D_x$;

    **else**

        Generate pseudo sentences $F_x$, $F_y$ by $\overleftarrow{\theta}$ , $\overrightarrow{\theta}$ on $D_y$, $D_x$;

        Weight $(F_x, D_y)$, $(F_y, D_x)$ by Eq.15;

    **end**

    Train $\overrightarrow{\theta}$ , $\overleftarrow{\theta}$ on synthetic data $(F_x, D_y)$, $(F_y, D_x)$; $Epoch = Epoch + 1$;

```
    end
return $\overrightarrow{\theta}, \overleftarrow{\theta}$ ;
```

## Training objective

In a typical machine-translation problem, given a source sentence x ∈ X, the goal is to find a high-scoring target sentence y ∈ Y; X, Y stand for the source space and target space. The score of each (x, y) pair is modeled by the probability that both sentences x and y will occur, denoted as P(x = x, y = y) (Lopez, 2008). If a perfect x → y translation system is available, P(x = x, y = y) = P(y = y|x = x)P(x = x).

Thus, we seek ˆθ; this is the optimal parameter for an x → y NMT system that estimates P(y|x) when:

$$P\left(X, Y; \hat{\theta}\right) = P(X, Y) \tag{4}$$

Where P(x, y; ˆθ) stands for the joint probability calculated by an NMT system using the parameter ˆθ. If only monolingual data are available in Y, ˆθ is difficult to calculate using only 4. Hence, we impose a necessary condition:

$$P\left(\mathbf{x}, \mathbf{y}; \hat{\theta})\right) = P(\mathbf{x}, \mathbf{y}) \rightarrow \sum_x P\left(\mathbf{x}, \mathbf{y}; \hat{\theta})\right) = \sum_x P(\mathbf{x}, \mathbf{y}) \leftrightarrow P\left(\mathbf{y}; \hat{\theta})\right) = P(\mathbf{y}) \tag{5}$$

This means that the marginal distribution expressed by ˆθ should be real when x is deemed as hidden variable. This is a compromise made to effectively train an unsupervised NMT system. As large amounts of monolingual data are available, it is possible to construct an LM that accurately estimates the real marginal distribution P(y). Thus, for the untutored θ values of an x → y NMT system, we deliberately narrow the gap between P(y) and P(y; θ):

$$\mathcal{L}^*(\theta) = -KL[P(Y)||P(Y;\theta)] \tag{6}$$

where KL[P(y)||P(y; θ)] is the Kullback–Leibler divergence between two distributions. Discarding irrelevant terms, the loss-maximizing 6 becomes:

$$\begin{aligned}
& \arg\max_\theta \mathcal{L}^*(\theta) \\
=\ & \arg\max_\theta \sum_y P(\mathbf{y}) \log P(\mathbf{y};\theta) \\
=\ & \arg\max_\theta \sum_y P(\mathbf{y}) \log \sum_x P(\mathbf{y}|\mathbf{x};\theta) P(\mathbf{x})
\end{aligned} \tag{7}$$

Note that Eq. 7 includes the unobserved data x and the logarithm of summation, which is difficult to calculate. Thus, we use the EM algorithm to train θ in an iterative manner. Consider the loss between iteration i + 1 and i.

$$\begin{aligned}
& \mathcal{L}^*(\theta^{i+1}) - \mathcal{L}^*(\theta^i) \\
=\ & \sum_y P(\mathbf{y}) \log \sum_x \frac{P(\mathbf{y}|\mathbf{x};\theta^{i+1})P(\mathbf{x})}{P(\mathbf{y};\theta^i)} \\
\geq\ & \sum_y P(\mathbf{y}) \sum_x P(\mathbf{x}|\mathbf{y};\theta^i) \log \frac{P(\mathbf{y}|\mathbf{x};\theta^{i+1})P(\mathbf{x})}{P(\mathbf{x}|\mathbf{y};\theta^i)P(\mathbf{y};\theta^i)}
\end{aligned} \tag{8}$$

Above, we apply Jensen's inequality when a y value is certain. The equality sign is valid when θ i+1 equals θ i . We define the evidence lower bound (ELBO) Hoffman et al. (2013) as:

$$ELBO(\theta, \theta^i) = \sum_y P(\mathbf{y}) \sum_x P(\mathbf{x}|\mathbf{y}; \theta^i) \log \frac{P(\mathbf{y}|\mathbf{x}; \theta)P(\mathbf{x})}{P(\mathbf{x}|\mathbf{y}; \theta^i)} \quad (9)$$

It is easy to show that L ∗ (θ) ≥ ELBO(θ, θi ) and L ∗ (θ i ) = ELBO(θ i , θi ). Thus, for an θ i+1 value satisfying ELBO(θ i+1, θi ) ≥ ELBO(θ i , θi ), we confirm:

$$\begin{aligned} \mathcal{L}^*(\theta^{i+1}) &\geq ELBO(\theta^{i+1}, \theta^i) \\ &\geq ELBO(\theta^i, \theta^i) \\ &= \mathcal{L}^*(\theta^i). \end{aligned} \quad (10)$$

Hence, we choose to maximize the ELBO of θ i+1; this is the M-step of the EM algorithm:

$$\theta^{i+1} = \arg\max_\theta ELBO(\theta, \theta^i) \quad (11)$$

We must calculate the following loss (this is the E-step of the EM algorithm):

$$\begin{aligned} \mathcal{L}(\theta, \theta^i) &= \sum_y P(\mathbf{y}) \sum_x P(\mathbf{x}|\mathbf{y}; \theta^i) \log P(\mathbf{y}|\mathbf{x}; \theta) \\ &= \mathbb{E}_{\mathbf{y} \sim P(\mathbf{y})} \left[ \mathbb{E}_{\mathbf{x} \sim P(\mathbf{x}|\mathbf{y}; \theta^i)} \log P(\mathbf{y}|\mathbf{x}; \theta) \right] \end{aligned} \quad (12)$$

A solution of Eq. 12 requires two sampling processes that generate training data for P(y|x; θ); this approach approximates the integral over the X and Y space. First, we randomly sample monolingual target sentences. The second sampling can proceed in two ways:

1) Use of the Bayes rules:

$$\begin{aligned} \mathbf{x}_{best} &= \arg\max_\mathbf{x} P(\mathbf{x}|\mathbf{y}; \theta^i) \\ &= \arg\max_\mathbf{x} P(\mathbf{y}|\mathbf{x}; \theta^i) P(\mathbf{x}) \end{aligned} \quad (13)$$

If a strong LM for language x is available, it is possible to sample natural sentences. However, it is necessary to use an encoder to choose all words of x because x serves as a condition. The initial state of the decoder is changed by each candidate word in x. The computational load is very high; it is impossible to perform the beam search of a typical neural encoder-decoder.

2) Alternatively, vanilla back-translation uses P(x|y; θ i back) directly, thus, not P(x|y; θ i ), to minimize the computational load; θ i back is a parameter of the backward model.This method slightly compromises mathematical soundness, and may generate noise Poncelas et al. (2018).

## Weighting of synthetic data

Given the high computational demand, and the noise issue, we combined the two methods mentioned above when engaging in the second sampling process. We heuristically leveraged the back-translation weights. Given the derived loss (Eq. 12), for each target sentence y, we tested all source sentences x, and assign them weights:

$$\begin{aligned} \mathcal{W}(\mathbf{x}, \mathbf{y}; \theta^i) &= P(\mathbf{y})P(\mathbf{x}|\mathbf{y}; \theta^i) \\ &= \frac{P(\mathbf{y})}{P(\mathbf{y}; \theta^i)} P(\mathbf{x}) P(\mathbf{y}|\mathbf{x}; \theta^i) \end{aligned} \quad (14)$$

Intuitively, strong LMs fine-tune the weights of synthetic data in two ways. For a target sentence y, if the current P(y; θ i ) is an overestimate of the probability, P(y)/P(y; θ i ) will be less than 1, and the weight of a sentence pair containing y will be reduced. If P(y; θ i ) is an underestimate of the probability, the weight will increase. Therefore, the modeled estimation inaccuracy of a target sentence y will be corrected. On the other hand, for a pseudo-source sentence x, P(x) is reliable when sampling, reducing the effects of unnatural sentences.

Thus, we applied weighting; we relaxed the synthetic data generated by back-translation. We proceeded as follows:

1) Treating the decoder as an LM, we used P(y; θ i dec) to estimate P(y; θ i ). θ i dec a decoder parameter. Thus, we chose a dummy as the source sentence; this avoids the need to calculate P(y; θ i ) = P x P(y|x; θ i )P(x). We followed (Ramachandran, Liu, and Le, 2016).

2) we normalize the logarithmic weight using the zeromean approach and then employed a sigmoid function to obtain the final weights:

$$\mathcal{W}^*(\mathbf{x}, \mathbf{y}; \theta^i) = sigmoid\left(ZM(\log \mathcal{W}(\mathbf{x}, \mathbf{y}; \theta^i))\right) \tag{15}$$

where ZM indicates zero-mean normalization. Sentence probability values are always separated by exponential gaps; training is dominated by a few highly weighted sentences. Normalization of logarithmic weights reduces the dominance of sentences with absolutely higher probabilities; the sigmoid function restricts the weights to within an appropriate interval (0, 1).

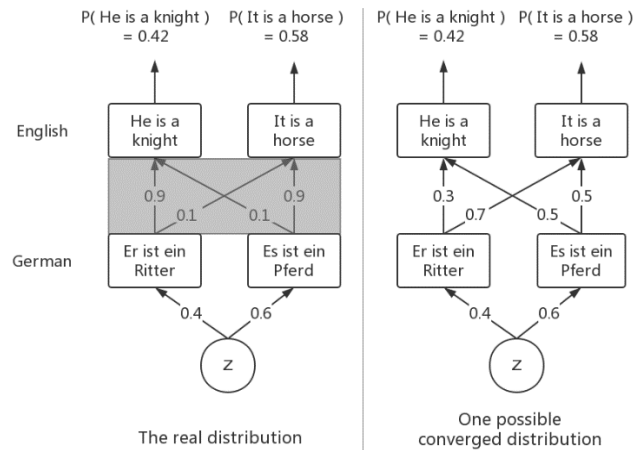

Figure 2: Illustration of the result when set goal as Eq. 5. Left denote the real distribution. Right is a possible training result, whose marginal distribution is equivalent to real, however posterior distribution does not converge to real due to the unobservability of real posterior distribution (indicated by shadow area). Fortunately, thanks to accuracy at marginal distribution, the result has been subjected to the solution space of a group of linear equations.

As Eq. 5 indicates, P(y; θ) = P(y) is only one necessary condition for attainment of fundamental goal Eq. 4. This simply constrains θ to a smaller space; θ satisfies:

$$W_\theta P_\mathbf{x} = P_\mathbf{y} \tag{16}$$

Where Wθ is a |Y| × |X| matrix defined by θ, denotes each P(y|x; θ). Px, Py is the probability vector of each language. ˆθ, the optimal parameters for an x → y NMT system, also satisfies Eq. 16. There is no guarantee that training of P(y|x; θ f inal) will converge to real distribution. Depending on θ 0 , the initial parameters, P(y|x; θ f inal) may converge relatively poorly (Figure 2). As the algorithm is sensitive to the initial value, it is important to carefully choose the initial parameters.

As Koehn and Knowles (2017) showed, SMT performs better than NMT in resource-poor environments. Thus, we used a temporary SMT to generate the initial synthetic data. Employing the "Noisy Channel" approach (Shannon, 1948), we used a well-trained LM and cross-lingual embedding to correct the word order and word-level translation; this is reminiscent of an unsupervised phrase-based SMT (PBSMT) (Lample et al., 2018).

Formally, we applied the Bayes rule:

$$\begin{aligned} \mathbf{x}_{best} &= \arg\max_{\mathbf{x}} P(\mathbf{x}|\mathbf{y}; \theta^0) \\ &= \arg\max_{\mathbf{x}} P(\mathbf{y}|\mathbf{x}; \theta^0) P(\mathbf{x}) \end{aligned} \quad (17)$$

Next, we employed a PBSMT(Zens, Och, and Ney, 2002) and Eq. 3 to decompose $P(y|x; \theta^0)$ into:

$$\begin{aligned} P(\mathbf{y}|\mathbf{x}; \theta^0) &= \prod_{i=1}^{n} \phi(y_i|x_i) \\ &= \prod_{i=0}^{n} \frac{\exp \lambda \cos <e_{x_i}, e_{y_i}>}{\sum_{y} \exp \lambda \cos <e_{x_i}, e_y>} \end{aligned} \quad (18)$$

in which y is segmented into a sequence of unrelated phrases y1, y2, ..., yn via inferred phrase table. We assumed that the probability distribution was uniform over all possible segmentations. Next, $P(y|x; \theta^0)$ was decomposed into a series of phrase translation probabilities φ(yi |xi) calculated using Eq. 3.

When applying the Bayes rule, the model may be perceived as log-linear in nature (Och and Ney, 2002), associated with certain artificial features such as grammatical rewards, unknown word and length penalties, and distortion scores. These complicate the issue. We added only an unknown word penalty and a distortion Koehn, Och, and Marcu (2003).

# Experiment

## Settings

We evaluated our method using three language pairs: English-French, English-German, and English-Russian. We used the BLEU score (Papineni et al., 2002) to assess translation quality.

**Dataset** All available sentences in the four languages available in NewsCrawl (a monolingual dataset of WMT) were used; these served as the baselines. We employed all available monolingual data when training the LMs. We randomly chose four monolingual source sentences per iteration for each language to generate synthetic data. The validation datasets were those of newstest2014(enfr) and newstest2016(en-de, en-ru); both include 3, 000 sentence pairs.

**Details** All data were tokenized and true-cased using Moses (Koehn et al., 2007) and segmented into subword symbols with the aid of Byte-Pair Encoding (BPE) (Sennrich, Haddow, and Birch, 2016); the shared vocabulary size was 60, 000. We use a kenLM (Heafield, 2011) and Fast-Text software Bojanowski et al. (2017) to generate word embeddings of dimension 512. Following Lample et al. (2018), we set λ of Eq. 18 to 30, and employed Vecmap[https://github.com/artetxem/vecmap] to perform cross-lingual embedding. When inferring initial data using the PBSMT, we implemented the Moses unknown word penalty and distortion score defaults. We did not further tune the PBSMT. For each language pair, we trained two independent NMT models (one in either translation direction) employing[https://github.com/tensorflow/tensor2tensor] , and we utilized beam searching (beam size 4) to generate subsequent synthetic data; testing featured a beam size of 16.

**Baseline** We compare our method to 7 baselines. The first baseline is supervised, train featured 0.6 million parallel sentences. The second baseline employs a twolanguage shared encoder based on a DAE (Artetxe et al., 2017). The third baseline features an additional adversarial training method(Lample et al., 2017). The fourth baseline introduces a weight- sharing mechanism to enhance performance (Yang et al., 2018).

The final baselines (5 to 7) are the strongest (Lample et al., 2018). Baseline 5 uses a DAE to substitute for and re-order synthetic data. Baseline 6 trains an unsupervised PBSMT system. Baseline 7 uses synthetic data generated by a PBSMT to tune the NMT further. Unlike our method, baseline 7 first fully trains an unsupervised PBSMT system, and then tunes an NMT system. We used the data generated by the initial PBSMT as initialization inputs.

# Results

Table 1 shows that our method allows our unsupervised NMT model to converge at higher BLEU scores (compared to those of prior baselines) in almost all of the six directions; performance is comparable to that of a supervised NMT model using 0.6 million parallel sentences. We performed several iterations; the detailed BLEU results are shown in Figure3. Further iterations afforded no additional improvements in BLEU scores. Some examples are shown in Table 3.

On the English-French task, the initial performance was better than those of the other two tasks, but subsequent training of the NMT model was not associated with immediate attainment of the strongest baseline. English and French constitute a strongly related language pair; it is simple to construct a strong phrase table. Thus, an SMT model performed better than an NMT model given an English-French task (Lample et al., 2018). As shown by the fifth baseline, our method renders an NMT model comparable to an SMT model in terms of an English-French task.

On the English-German and English-Russian tasks, our model significantly outperformed the previous best models by about 3 BLEU; thus, we define a new state-of-the-art standard. By appropriately weighting the synthetic data, and optimizing initialization, an NMT system can be guided in the correct direction.

Notably, the BLEUs of systems that focused on English always increased to the interval, suggesting that training advances in a manner whereby forward model enhancement relies on the performance of the backward model.

| Method | fr-en | en-fr | de-en | en-de | ru-en | en-ru |
|---|---|---|---|---|---|---|
| Supervised(0.6 million) | 28.87 | 29.45 | 28.24 | 23.35 | - | - |
| (Artetxe et al., 2017) | 15.56 | 15.13 | 10.21 | 6.89 | - | - |
| (Lample et al., 2017) | 14.31 | 15.05 | 13.33 | 9.64 | - | - |
| (Yang et al., 2018) | 15.58 | 16.97 | 14.62 | 10.86 | - | - |
| (Lample et al., 2018), NMT | 24.18 | 25.14 | 21.00 | 17.16 | 9.09 | 7.98 |
| (Lample et al., 2018), PBSMT | 27.16 | 28.11 | 22.68 | 17.77 | 16.62 | 13.37 |
| (Lample et al., 2018), PBSMT+NMT | **27.68** | 27.60 | 25.19 | 20.23 | 16.62 | 13.76 |
| **Our method** | 27.49 | **28.22** | **28.92** | **23.61** | **19.57** | **16.14** |

Table 1: Comparison with previous work. Beam size is set to 16.

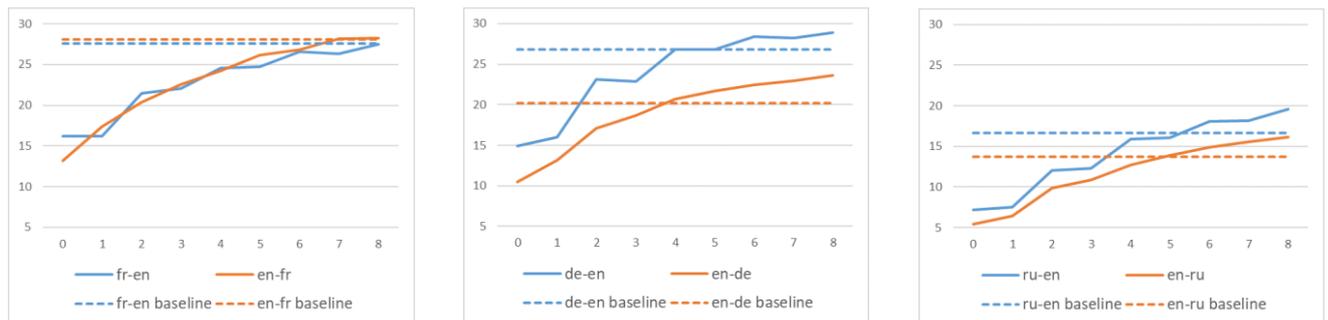

(a) English-French.  (b) English-German.  (c) English-Russian.

Figure 3: Bleu scores each iterations. Beam size is set to 16. Baselines are each strongest baseline of each translation task in table1

# Ablation Study

| Model | de-en | en-de |
|---|---|---|
| Full model | 28.87 | 23.61 |
| Without weighting | 26.09 | 19.83 |
| word-by-word initialization | 17.19 | 14.62 |

Table 2: Ablation study on English-German task.

We tested: 1) removal of weighting; and 2) word-by-word initialization (Table 2). When the weightings of synthetic data were removed, the scores were similar to those of the seventh baseline, perhaps because an auxiliary SMT system was in play. We found that use of an SMT system to generate the initial synthetic data was equivalent to employment of a fully trained SMT system to fine-tune the NMT. The third line of Table 2 stresses the need for appropriate initialization.

## Case Study

| | |
|---|---|
| Source | er argumentiert , dass er zunehmend von kommerziellen Rivalen nicht zu unterscheiden ist . |
| Ref | he argues that he is increasingly indistinguishable from commercial rivals. |
| Initial | he argued that he increasingly by commercial rivals not to distinguish is . " |
| Iter 1 | he also argues that he is increasingly of commercial rivals not to differentiate . |
| Iter 4 | he argues that he is increasingly not going to distinguish from commercial rivals . |
| Iter 7 | he also argued that he is increasingly unable to distinguish from any commercial rival . |
| Source | " des stratégies pédagogiques différentes , c' est ça le véritable besoin " , résume-t-elle . |
| Ref | " the real need is for different educational strategies , " she summarises . |
| Initial | " the educational strategies , it is the ultimate " it needs , " and various |
| Iter 1 | " the educational strategies , different ones , it 's something the ultimate need . " |
| Iter 4 | " from different teaching strategies , this is really the ultimate need , " he writes . |
| Iter 7 | " different teaching strategies is just the ultimate need , " she say . |

Table 3: Selected test cases.

| Synthetic Source | Monolingual Target | Weight |
|---|---|---|
| but she didn't pick the **small** computers . | mitnehmen durfte sie den kleinen Rechner aber nicht . | 0.723 |
| but she didn't pick the **low** computers . | mitnehmen durfte sie den kleinen Rechner aber nicht . | 0.688 |
| families with children are also in hotels with the disclaimer comfort hotel the exception. | Familien mit Kindern sind dagegen in Hotels mit dem Zusatz Wohlfühlhotel die Ausnahme. | 0.549 |

Table 4: Selected training cases and their weight in English-German direction, iteration 4. In second case, word <small> is manually replaced with <low> .

To understand more fully how the score function (Eq. 15) corrects training of the NMT model, we illustrate three cases in Table 4. We manually replaced "small" by "low" in the first case; this is inappropriate, and the weight declines on LM scanning using the current NMT model. Moreover, given various synthetic sentence pairs, the model will find the more helpful cases and increase their weights, as may be seen by comparing the first and third cases. Generally, the model will prioritize frequently occurring sentences (such as short sentences); these are usually easier to translate. More complex and less common sentences will receive lower weights.

# Related Work

NMTs that must operate in extremely resource-poor conditions are of great interest. Given the limited supervision, several efforts have been made to boost NMT systems using monolingual data, principally by leveraging bilingual lexica (Klementiev et al., 2012), by employing language models (Ramachandran, Liu, and Le, 2016; He et al., 2016; Gulcehre et al., 2015), and by exploiting iterative back-translation (Sennrich, Haddow, and Birch, 2015; Zhang et al., 2018; Hoang et al., 2018).

Following the pioneering work of Ravi and Knight (2011), some authors have attempted to create unsupervised NMTs (Artetxe et al., 2017; Yang et al., 2018; Lample et al., 2017, 2018). In such works, source sentences are viewed as internal information and are mapped into a latent space that is not relevant to the language per se; target sentences are generated via DAE (Vincent et al., 2008). Back-translation was employed in almost all previous works. Lample et al.

(2018) further tuned an NMT model using data generated by a PBSMT; performance improved significantly.

Similar to our studies, some authors have sought to improve the initialization parameters of NMT models using weight-generated corpora during back-translation. Ramachandran, Liu, and Le (2016) initialized both the encoder and decoder as LMs. Encouraged by the success of bilingual lexicon induction (Fung and Yee, 1998; Conneau et al., 2017; Artetxe, Labaka, and Agirre, 2018), cross-lingual embedding is now widely used to initialize unsupervised models (Artetxe et al., 2017; Yang et al., 2018; Lample et al., 2017, 2018). Zhang et al. (2018) weighted synthetic data by translation probabilities computed with the aid of a backward model. He et al. (2016) viewed the LM scores as rewards of a reinforcement learning framework.

# Conclusion

We sought to improve the performance of unsupervised NMT models. We employed an LM and an inferred bilingual dictionary to construct a PBSMT system, and initialized the NMT model using the PBSMT-generated data. We then employed non-vanilla back-translation to formulate a weight function for synthetic data; this allowed the NMT model to perform better than before. We applied our method to analysis of three language pairs; we have established new state-of-the-art performance parameters for unsupervised machine translation

# References

Artetxe, M.; Labaka, G.; Agirre, E.; and Cho, K. 2017. Unsupervised neural machine translation. *arXiv preprint arXiv:1710.11041*.

Artetxe, M.; Labaka, G.; and Agirre, E. 2018. Gen- eralizing and improving bilingual word embedding mappings with a multi-step framework of linear trans- formations. In *Thirty-Second AAAI Conference on Artificial Intelligence*.

Bahdanau, D.; Cho, K.; and Bengio, Y. 2014. Neural machine translation by jointly learning to align and translate. *arXiv preprint arXiv:1409.0473*.

Bojanowski, P.; Grave, E.; Joulin, A.; and Mikolov, T. 2017. Enriching word vectors with subword informa- tion. *Transactions of the Association for Computa- tional Linguistics* 5(1):135–146.

Cho, K.; Van Merriënboer, B.; Gulcehre, C.; Bahdanau, D.; Bougares, F.; Schwenk, H.; and Bengio, Y. 2014. Learning phrase representations using rnn encoder- decoder for statistical machine translation. *arXiv preprint arXiv:1406.1078*.

Conneau, A.; Lample, G.; Ranzato, M.; Denoyer, L.; and Jégou, H. 2017. Word translation without parallel data. *arXiv preprint arXiv:1710.04087*.

Fung, P., and Yee, L. Y. 1998. An ir approach for translating new words from nonparallel, comparable texts. In *36th Annual Meeting of the Association for Computational Linguistics and 17th International Conference on Computational Linguistics, Volume 1*, volume 1.

Gulcehre, C.; Firat, O.; Xu, K.; Cho, K.; Barrault, L.; Lin, H.-C.; Bougares, F.; Schwenk, H.; and Bengio, Y. 2015. On using monolingual corpora in neural machine translation. *arXiv preprint arXiv:1503.03535*.

He, D.; Xia, Y.; Qin, T.; Wang, L.; Yu, N.; Liu, T.-Y.; and Ma, W.-Y. 2016. Dual learning for machine trans- lation. In *Advances in Neural Information Processing Systems*, 820–828.

Heafield, K. 2011. Kenlm: Faster and smaller language model queries. In *In Proc. of the Sixth Workshop on Statistical Machine Translation*.

Hoang, V. C. D.; Koehn, P.; Haffari, G.; and Cohn, T. 2018. Iterative back-translation for neural machine translation. In *Proceedings of the 2nd Workshop on Neural Machine Translation and Generation*, 18–24. Melbourne, Australia: Association for Computational Linguistics.

Hoffman, M. D.; Blei, D. M.; Wang, C.; and Paisley, J. 2013. Stochastic variational inference. *The Journal of Machine Learning Research* 14(1):1303–1347.

Klementiev, A.; Irvine, A.; Callison-Burch, C.; and Yarowsky, D. 2012. Toward statistical machine trans- lation without parallel corpora. In *Proceedings of the 13th Conference of the European Chapter of the Association for Computational Linguistics*, 130–140. Association for Computational Linguistics.

Koehn, P., and Knowles, R. 2017. Six challenges for neural machine translation. *arXiv preprint arXiv:1706.03872*.

Koehn, P.; Hoang, H.; Birch, A.; Callison-Burch, C.; Federico, M.; Bertoldi, N.; Cowan, B.; Shen, W.; Moran, C.; Zens, R.; et al. 2007. Moses: Open source toolkit for statistical machine translation. In *Proceedings of the 45th annual meeting of the associ- ation for computational linguistics companion volume proceedings of the demo and poster sessions*, 177–180.

Koehn, P.; Och, F. J.; and Marcu, D. 2003. Statistical phrase-based translation. In *Proceedings of the 2003 Conference of the North American Chapter of the Association for Computational Linguistics on Human Language Technology-Volume 1*, 48–54. Association for Computational Linguistics.

Lample, G.; Conneau, A.; Denoyer, L.; and Ran- zato, M. 2017. Unsupervised machine translation using monolingual corpora only. *arXiv preprint arXiv:1711.00043*.

Lample, G.; Ott, M.; Conneau, A.; Denoyer, L.; and Ranzato, M. 2018. Phrase-based & neural unsupervised machine translation. *arXiv preprint arXiv:1804.07755*.

Lopez, A. 2008. Statistical machine translation. *ACM Comput. Surv.* 40(3):8:1–8:49.

Och, F. J., and Ney, H. 2002. Discriminative training and maximum entropy models for statistical machine translation. In *Proceedings of the 40th annual meeting on association for computational linguistics*, 295–302. Association for Computational Linguistics.

Papineni, K.; Roukos, S.; Ward, T.; and Zhu, W.-J. 2002. Bleu: a method for automatic evaluation of machine translation. In *Proceedings of the 40th annual meeting on association for computational linguistics*, 311–318. Association for Computational Linguistics.

Poncelas, A.; Shterionov, D.; Way, A.; Wenniger, G. M. d. B.; and Passban, P. 2018. Investigating backtrans- lation in neural machine translation. *arXiv preprint arXiv:1804.06189*.

Ramachandran, P.; Liu, P. J.; and Le, Q. V. 2016. Unsupervised pretraining for sequence to sequence learning. *arXiv preprint arXiv:1611.02683*.

Ravi, S., and Knight, K. 2011. Deciphering foreign language. In *Proceedings of the 49th Annual Meet- ing of the Association for Computational Linguistics: Human Language Technologies*, 12–21.

Sennrich, R.; Haddow, B.; and Birch, A. 2015. Improv- ing neural machine translation models with monolin- gual data. *arXiv preprint arXiv:1511.06709*.

Sennrich, R.; Haddow, B.; and Birch, A. 2016. Neu- ral machine translation of rare words with subword units. In *Proceedings of the 54th Annual Meeting of the Association for Computational Linguistics (Vol- ume 1: Long Papers)*, 1715–1725. Berlin, Germany: Association for Computational Linguistics.

Shannon, C. E. 1948. A mathematical theory of commu- nication. *Bell system technical journal* 27(3):379–423.

Sutskever, I.; Vinyals, O.; and Le, Q. V. 2014. Se- quence to sequence learning with neural networks. In *Advances in neural information processing systems*, 3104–3112.

Vaswani, A.; Shazeer, N.; Parmar, N.; Uszkoreit, J.; Jones, L.; Gomez, A. N.; Kaiser, Ł.; and Polosukhin, I. 2017. Attention is all you need. In *Advances in Neural Information Processing Systems*, 5998–6008.

Vincent, P.; Larochelle, H.; Bengio, Y.; and Manzagol, P.-A. 2008. Extracting and composing robust features with denoising autoencoders. In *Proceedings of the 25th international conference on Machine learning*, 1096–1103. ACM.

Yang, Z.; Chen, W.; Wang, F.; and Xu, B. 2018. Un- supervised neural machine translation with weight sharing. *CoRR* abs/1804.09057.

Zens, R.; Och, F. J.; and Ney, H. 2002. Phrase-based statistical machine translation. In *Annual Conference on Artificial Intelligence*, 18–32. Springer.

Zhang, Z.; Liu, S.; Li, M.; Zhou, M.; and Chen, E. 2018. Joint training for neural machine translation models with monolingual data. In *Thirty-Second AAAI Conference on Artificial Intelligence*.